\documentclass[10pt,twocolumn,letterpaper]{article}

\usepackage{cvpr}              

\usepackage{times}
\usepackage{epsfig}
\usepackage{graphicx}
\usepackage{amsmath}
\usepackage{amssymb}
\usepackage{capt-of}
\usepackage{multirow}
\usepackage{algorithm}
\usepackage{algorithmic}
\usepackage{caption}
\usepackage{bbm}

\usepackage[accsupp]{axessibility} 
\newcommand*\samethanks[1][\value{footnote}]{\footnotemark[#1]}

\usepackage[pagebackref,breaklinks,colorlinks]{hyperref}

\usepackage[capitalize]{cleveref}
\crefname{section}{Sec.}{Secs.}
\Crefname{section}{Section}{Sections}
\Crefname{table}{Table}{Tables}
\crefname{table}{Tab.}{Tabs.}

\def\cvprPaperID{11994} 
\def\confName{CVPR}
\def\confYear{2022}

\begin{document}

\title{Local-Adaptive Face Recognition via Graph-based Meta-Clustering and Regularized Adaptation}

\author{
    Wenbin Zhu\textsuperscript{1\thanks{~indicates equal contribution.}} \qquad 
    Chien-Yi Wang\textsuperscript{2\samethanks} \qquad 
    Kuan-Lun Tseng\textsuperscript{2} \qquad 
    Shang-Hong Lai\textsuperscript{2} \qquad
    Baoyuan Wang\textsuperscript{3} \qquad
    \\
    \textsuperscript{1}Microsoft Cloud and AI
    \quad
    \textsuperscript{2}Microsoft AI R\&D Center, Taiwan
    \quad
    \textsuperscript{3}Xiaobing.AI
    \\
    {\tt\small \{wenzh, chiwa, kutseng, shlai\}@microsoft.com}
    \quad
    {\tt\small zjuwby@gmail.com}
} 
\maketitle

\begin{abstract}
   Due to the rising concern of data privacy, it's reasonable to assume the local client data can't be transferred to a centralized server, nor their associated identity label is provided. To support continuous learning and fill the last-mile quality gap,  we introduce a new problem setup called ``local-adaptive face recognition (LaFR)". Leveraging the environment-specific local data after the deployment of the initial global model, LaFR aims at getting optimal performance by training local-adapted models automatically and un-supervisely, as opposed to fixing their initial global model. We achieve this by a newly proposed embedding cluster model based on Graph Convolution Network (GCN), which is trained via meta-optimization procedure. Compared with previous works, our meta-clustering model can generalize well in unseen local environments. With the pseudo identity labels from the clustering results, we further introduce novel regularization techniques to improve the model adaptation performance. Extensive experiments on racial and internal sensor adaptation demonstrate that our proposed solution is more effective for adapting face recognition models in each specific environment. Meanwhile, we show that LaFR can further improve the global model by a simple federated aggregation over the updated local models.
\end{abstract}

\begin{figure}[t!]
    \centering
    \includegraphics[width=.9\linewidth]{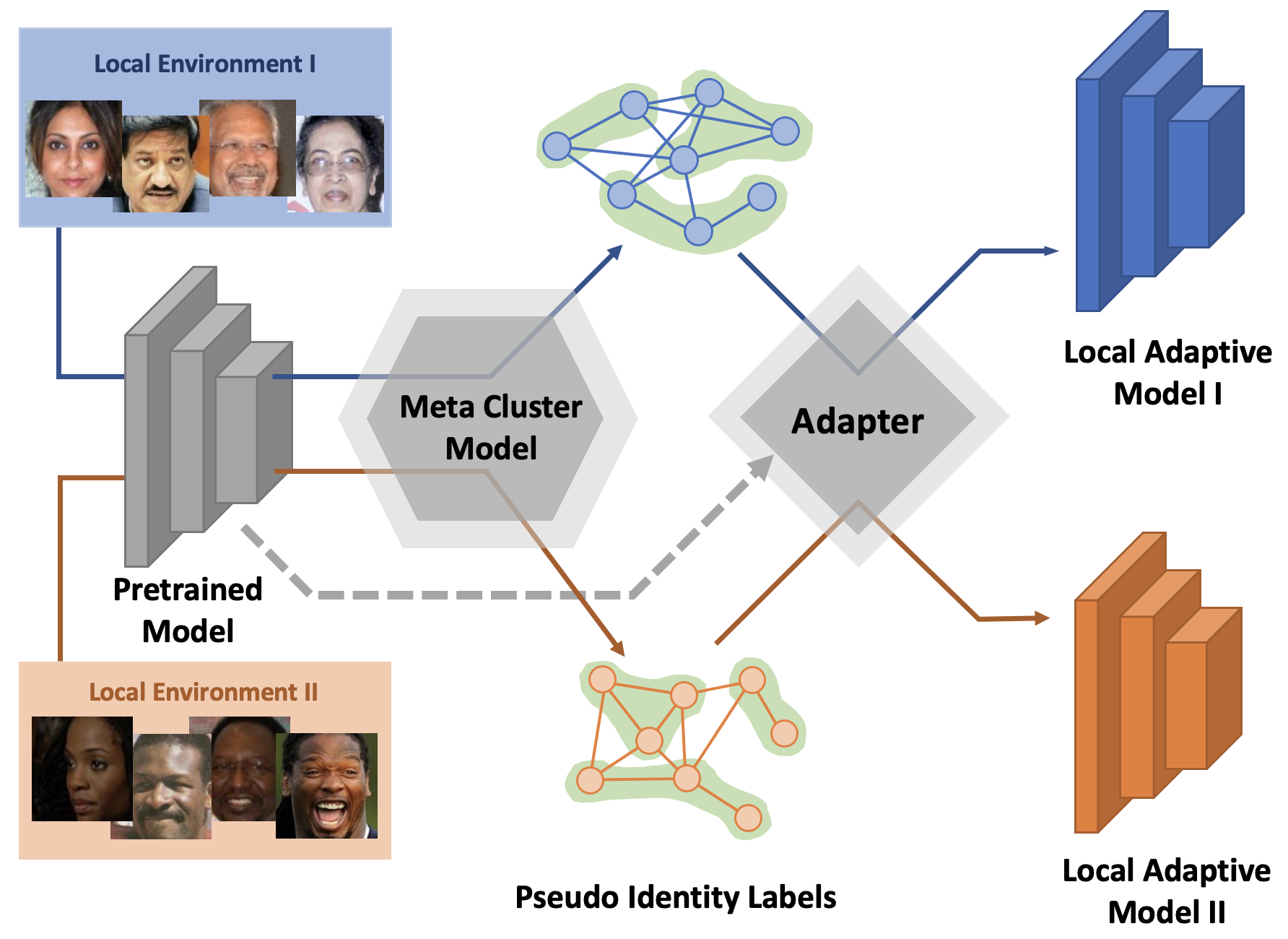}
    \caption{Local-Adaptive Face Recognition (LaFR): For each local environment, a specialized model is produced by the adapter module with only the pre-trained model and images from the environment. Note that there is no real identity label associated with the images, the meta-cluster model generates pseudo labels for robust model adaptation.}
    \label{fig:problem_setting}
\vspace{-4.5mm}
\end{figure}

\section{Introduction}\label{sec.intro}
Face recognition~\cite{Deepface14} has been commercialized widely for a variety of applications, such as FaceID, surveillance monitoring. The COVID-19 pandemic even accelerates the biometric technologies for touch-less solutions, such as face recognition enabled payment and access control. Although remarkable progress has been achieved lately, one has to admit that face recognition still hasn't been fully solved. Among many other remaining challenges (i.e., vulnerability for adversarial attack~\cite{Dong_2019_CVPR_attack}), how to scale up the representation learning to reduce the risk of fairness and bias to support various local environments becomes a more urgent challenge. As studied in previous works \cite{Wang_2019_ICCV, Yin_2019_CVPR, Devil17}, such fairness and bias issues come from both algorithmic design and under-represented data distributions. For example, when the model is predominantly trained on RGB images, it generalizes poorly for images captured by Infrared cameras. Likewise, for a model pre-trained on Caucasian only, it performs substantially worse for African and Indian.

While it is worthwhile to push the domain-invariant face recognition \cite{Guo_2020_CVPR, Shi_2020_CVPR} with the hope of generalizing to everywhere without adaptation, it is arguably that the challenges for real-world scenarios could be more than we expected. So the question is: given an imperfect pre-trained model, how can we improve and fill the last-mile performance gap for each local environment and thereafter scale up the process? In this paper, we are interested in studying how to properly adapt the pre-trained model to a ``specialized" one that tailors for the specific environment in an automatic and unsupervised manner. Here, the ``environment" could be defined broadly, including a specific new camera sensor (i.e. an infrared camera with particular wavelength), a unique identity distribution with racial bias, or a physical environment that has unique camera placement and lighting condition, etc. We call such a problem setup as ``Local-Adaptive Face Recognition (LaFR)",
whose workflow (see Fig.~\ref{fig:problem_setting}) starts from an imperfect pre-trained global model deployed to a specific environment, where it accumulates some amount of new data. It then applies unsupervised adaptation technique to adapt the initial model locally, hence no data is transferred to server. Finally, after the adaptation, the new model is expected to perform much better than the initial global model as it is trained to tailor that environment. As an optional step, Federated Learning \cite{FederatedAVg} is further employed to aggregate many such local models in a secure manner. Therefore, ``LaFR" essentially provides a way to scale up the representation learning and model generalization via such ``dual-loop" paradigm.

Although unsupervised domain adaptation (UDA) \cite{UDA} has been widely studied in person re-identification (re-ID)~\cite{song2018unsupervised,ssg_reID_2019,Lin_Dong_Zheng_Yan_Yang_2019,Zhuang_2020_MM,TIP2020_ReID_Lin,kumar2020unsupervised,ye2021deep}, it is much less explored in face recognition except \cite{Wang_2019_ICCV,Eric17}. Most of those works either designed special for person re-ID~\cite{ssg_reID_2019}, or their setups require both source and target dataset to be available during the adaptation stage, or they only aim at closed-set problem. Moreover, person re-ID works heavily depend on variants of triplet loss, as we show in prior works, there are more robust losses (i.e., CircleLoss~\cite{CircleLoss}) that proved to work better for face recognition.

To overcome the challenges, we first introduce a graph-based meta-clustering algorithm designed to predict pseudo labels for any unlabelled dataset.  To do this, we collect a set of labeled datasets from multiple domains and extract their face embeddings from the given pre-trained model, we then apply Graph Convolution Network (GCN) to model the non-convex structure relationship for face embeddings within each set, which is trained efficiently through meta-learning \cite{MAML} to make the cluster prediction more generalizable for the unseen dataset. Secondly, to better facilitate the transfer learning, we introduce a new technique by transferring the representation of class (pseudo label) center from the pre-trained model to the classifier of the new model and keep it fixed while only fine-tuning the feature representation in the context of margin-based training objectives such as \cite{Wang_2018_amsoftmax, ArcFace19, CircleLoss}. Moreover, instead of regularizing the feature distance (commonly used in knowledge distillation \cite{LRFR19, hinton2015distilling, Sohn_2017_ICCV}), we regularize the network weights to ensure a small deviation between the pre-trained and the new local model. 
    
To summarize, we make the following major contributions: (1) we introduce a novel unsupervised model adaptation problem setup for face recognition, we argue that it's practical yet scalable motivated from both continuous learning and data privacy concerns; (2) we use graph convolution network (GCN) to model the dataset structure and predict the clustering labels, through a meta learning framework; (3) Our novel regularized center transfer (RCT) technique can significantly reduce the risk of overfitting and improve transfer learning performance for even smaller datasets; (4) Experiments show that our entire solution not only outperform other strong baselines for local adaption but also enable the federated learning to further improve the global model.

\begin{figure*}
    \centering
    \includegraphics[width=.99\linewidth]{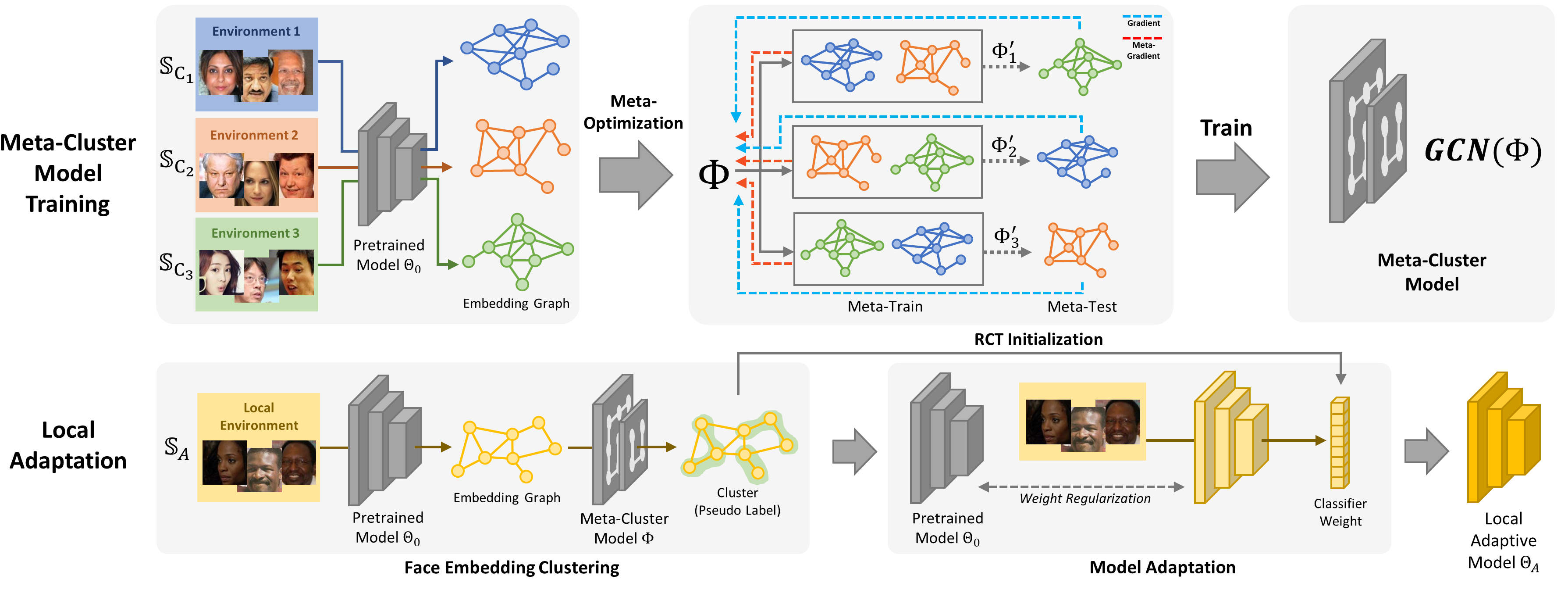}
    \caption{Overview of our Local-Adaptive face recognition (LaFR) framework. \textbf{Meta-Cluster Model Training:} A Graph Convolution Network (GCN) based embedding cluster model is trained with the meta-optimization procedure to generalize well in unseen local environments. \textbf{Test-time Local Adaptation:} In the local adaptation phase, we obtain the pseudo identity label of local face images from the meta-cluster model. Given the pre-trained face recognition model, local images, and their corresponding pseudo labels, the proposed Regularized Center transfer (RCT) technique can adapt the model more robustly to produce a specialized face recognition model.}
    \label{fig:overview}
\end{figure*}
\section{Related Works}
\paragraph{Generalized Face Recognition}
Most of the recent works focus on generalizing the representation power through novel loss functions, such as CircleLoss~\cite{CircleLoss}, Arcface~\cite{ArcFace19}, Regularface~\cite{Duan_2019_CVPR} and AM-SoftMax \cite{Wang_2018_amsoftmax}, with both solid theoretical foundation and remarkable empirical results are reported in SOTA models. A set of other new works target to improve the performance from the data distribution perspective, including data augmentation \cite{Yin_2019_CVPR} for under-represented classes, improved training strategy for unbalanced and long-tailed data \cite{Zhong_2019_CVPR}, and uncertainty modeling for noisy training set \cite{Chang_2020_CVPR}. More recently, generalized representation learning has received more and more attention. \cite{shi2020towards} learns universal representations to tackle pose, resolution, and occlusion variation, while~\cite{Guo_2020_CVPR} uses meta-learning to make it easily generalized to the unseen domain without requiring adaptation. \cite{Wang_2019_CVPR} proposes to use the adversarial decorrelation technique to make the identity representation invariant to age information. Causal relationship modeling through Invariant Risk Minimization~\cite{IRM} is a promising direction to address the out-of-distribution (OOD) challenge, yet, no impressive results have been reported so far for face recognition. Our method is both compatible and complementary with all those latest works. Improving the generalization capability of the pre-trained model will provide a better initialization for subsequent adaptation.

\paragraph{Unsupervised Domain Adaptation (UDA)} There is a large body of research works in unsupervised domain adaption~\cite{UDA} for person re-identification~\cite{song2018unsupervised,ssg_reID_2019,Lin_Dong_Zheng_Yan_Yang_2019,Zhuang_2020_MM,TIP2020_ReID_Lin,kumar2020unsupervised,ye2021deep}, semantic segmentation~\cite{UDA_Segmentation2020}, image classifications~\cite{wang2021unsupervised,Eric17,Mingsheng15} and attribute recognition~\cite{ji2020unsupervised}. However, most of those approaches either target for closed-set setting or require both source and target to be available during the adaptation to align the feature distribution through maximum mean discrepancy (MMD)~\cite{tolstikhin2016minimax}. Conceptually, a clustering technique is first needed to assign a pseudo label to each image before the training. Prior person re-ID works have explored various heuristic methods such as bottom-up cluster~\cite{Lin_Dong_Zheng_Yan_Yang_2019} and tracklet-based clustering~\cite{kumar2020unsupervised}. In contrast, our clustering is achieved through a graph convolution network (GCN) to learn the intrinsic structure for each unseen data-set via meta-learning. To our best knowledge, there is little work of applying UDA for face recognition \cite{UDA_FG18, Sohn_2017_ICCV}. Perhaps the most closely related work is \cite{Wang_2019_ICCV} in which a traditional MMD loss is required to be employed. However, as we argued before, in the context of ``LaFR", the source data is not available during the adaptation due to the data privacy and security concern.




\paragraph{Graph Convolution Network based Face Clustering}
Face clustering is another fundamental problem in computer vision, which is also quite related to our work. A recent series of studies \cite{yang2019learning, yang2020learning, wang2019gncclust} show that supervised clustering performs significantly better than the conventional unsupervised ones such as K-Means~\cite{Kmeans} and Spectral-Clustering~\cite{Spectral_clustering}. The reason behind is that these Graph Convolution Network (GCN)~\cite{kipf2017semisupervised} based methods are more capable of finding the local non-convex structures. Inspired by such remarkable progress, we move one step further and marry GCN-based clustering with meta-learning~\cite{MAML} to make the resulting clustering model more adaptive to the unseen dataset.






\section{Overview}
Given an imperfect pre-trained model $\Theta_0$, our whole system requires a labeled dataset $\mathbb{S}_{C}$ in the global-end, which consists of a few subsets ($\mathbb{S}_{{C}_1},...,\mathbb{S}_{{C}_k}$) from multiple domains to train a novel graph-based meta-clustering model $\Phi$. Once trained, given any unlabelled new dataset $\mathbb{S}_\mathcal{A}$ on a specific local environment, we will feed it into $\Phi$ to obtain the pseudo label for each image. Then, we apply our proposed regularized center transfer (RCT) technique to perform the adaptation and produce a specialized local model $\Theta_{A}$ tailors for that environment. Figure \ref{fig:overview} illustrates the framework of our system.
\section{Graph-based Meta-Clustering Learning}
\subsection{Preliminary of Graph Convolution Network}
Given a face embedding set $\mathbb{S} = \{{f_{i}, y_{i}}\}^{N}_{i=1}$, which $y_{i}$ is the corresponding identity label of each embedding. We first define a graph $\mathcal{G}(\mathcal{V},\mathcal{E})$ that connects each image with its K-NN neighbor, and the affinity matrix $\mathcal{A}$ is computed by cosine similarity between $f_i$ and its neighbor $f_j(j\in \mathcal{N}_i )$, so  $\mathcal{A}_{ij}=a_{ij} = \cos(f_i,f_j)$  and $\mathcal{N}_i$ represents the neighbors of $f_i$. Let us use $F_0 \in \mathbb{R}^{N \times d_{in}}$ to represent the input feature embedding for a Graph Convolution Network(GCN), then after $l$ layers of convolutions on the graph, the feature embedding $F_{l} \in \mathbb{R}^{N \times d_{out}}$ becomes

\begin{equation}
    F_{l} = \sigma(g(\Bar{A},F_{l-1})W^{\mathcal{G}}_{l-1})
\end{equation}
where $\Bar{A} = \Bar{D}^{-1}(A+I)$ and $\Bar{D}_{ii} = \sum_j(A+I)_j$ is a diagonal degree matrix, $W^{\mathcal{G}}_{l-1} \in \mathbb{R}^{2 \times d_{in} \times d_{out}}$ is a trainable matrix to transfer the input embedding into a new space, $\sigma$ is the nonlinear activation function(i.e., ReLU), and $g(\cdot,\cdot)$ is a concatenation function following previous works \cite{yang2019learning,yang2020learning,wang2019gncclust}, which is defined as 
\begin{equation}
    g(\Bar{A}, F_{l-1}) = [(F_{l-1})^T, (\Bar{A}F_{l-1})^T]^T
\end{equation}

In the context of this work, as we want to learn how to predict the face clustering for a given dataset, we let the last layer output a 1-D confidence value for each vertex. Assume $c'$ represents the vector of confidence values for each vertex, then
\begin{equation}
    c' = F_lW^{\mathcal{G}} +b
\end{equation}
where $W^{\mathcal{G}}$ is the  projection matrix and $b$ is the bias, both are learnable. Again, following the design of ``GCN-V" \cite{yang2020learning}, we define the ground-truth confidence $c_i$ as 
\begin{equation}
    c_i = \frac{1}{|\mathcal{N}_i|} \sum_{j \in \mathcal{N}_i} (\mathbbm{1}(y_j=y_i) - (\mathbbm{1}(y_j \neq y_i)))\cdot a_{ij} 
\end{equation}
where $\mathbbm{1}()$ is an indicator function. Therefore, we minimize the following loss to train GCN:

\begin{equation}
    \mathcal{L}^{\mathcal{G}}(\Phi) = \frac{1}{N} \sum_i^N |c_i - c'_i|
\end{equation}
where $c'_i$ is the confidence value from $c'$ for the corresponding vertex $i$, and $\Phi$ represents the parameters of the GCN model. Intuitively, a lower $c_i$ means this vertex could lie near the boundary between two different clusters according to the local graph structure.

\subsection{Meta-Clustering with GCN}
By no means we are the first to apply GCN for face clustering given the impressive works done recently \cite{yang2019learning, yang2020learning, wang2019gncclust}. Nevertheless, our work is compatible yet builds on top of them by changing the training strategy into a meta-learning paradigm. We are interested in solving the aforementioned Local-Adaptive face recognition scenarios, where the pre-trained model is already deployed and their source data is not available anymore. Therefore, an automatic and unsupervised adaptation from a pre-trained model becomes a necessary and urgent problem, in order to massively scale-up the adaptation. 

To train a meta-clustering model on GCN, we first collect a few labeled datasets $\mathbb{S}_{C_1},...,\mathbb{S}_{C_k}$ from $C_K$ different domains and then perform domain-level sampling for outer-loop iteration, where we sample $C_K-1$ domains data for the meta-train while the data from the remaining domain for the meta-test. Our goal is to let this GCN be more generalizable for any dataset from unseen domains. We describe the detailed training procedure in Algorithm \ref{alg:meta-clustering}. Specifically, We start with a randomly initialized GCN parameter setting $\Phi_0$, and for each meta-train iteration, we perform the conventional GCN training via the following equation:

\begin{equation}
    \Phi^{'} = \Phi - \alpha \nabla_{\Phi} \mathcal{L}_{mtr}^{\mathcal{G}}(\Phi)
\end{equation}
where $\mathcal{L}_{mtr}^{\mathcal{G}}(\Phi)$ denotes the loss on the meta-train dataset.
The model is then tested on the meta-test dataset. Similarly, we compute the corresponding loss as $\mathcal{L}_{mte}^{\mathcal{G}}(\Phi^{'})$ with the updated model $\Phi^{'}$. The meta-model will then be updated jointly from both gradients (Line $\#$8). This process is iterated until it reaches the maximum.

\begin{algorithm}[!tb]
\caption{Graph-based Meta-Clustering}
\label{alg:meta-clustering}
\begin{algorithmic}[1]
\REQUIRE 
 Initialize $\Phi$ as $\Phi_0$;
 Datasets $\mathbb{S}$[$\mathbb{S}_{C_1},...,\mathbb{S}_{C_k}$]; 
 Hyper-parameters $\alpha,\beta,\xi$
\ENSURE $\Phi$\\
\FOR{ iter $<$ MaxIter} 
\STATE \textbf{Split:} $\mathbb{S}$ into $\mathbb{S}_{mtr}$ and $\mathbb{S}_{mte}$ randomly
\STATE \textbf{Meta-Train:}
\STATE Train on $\mathbb{S}_{mtr}:$ $\Phi^{'} = \Phi - \alpha \nabla_{\Phi} \mathcal{L}_{mtr}^{\mathcal{G}}(\Phi)$ 
\STATE \textbf{Meta-Test:}
\STATE Compute loss on $\mathbb{S}_{mte}:\mathcal{L}_{mte}^{\mathcal{G}}(\Phi^{'})$ 
\STATE \textbf{Meta-optimization:} Update $\Phi$
\STATE $\Phi = \Phi - \beta(\nabla_{\Phi} \mathcal{L}_{mtr}^{\mathcal{G}}(\Phi) + \xi \nabla_{\Phi} \mathcal{L}_{mte}^{\mathcal{G}}(\Phi^{'}))$ 
\ENDFOR
\end{algorithmic}
\end{algorithm}
\section{Unsupervised Model Adaptation}
As discussed before, our pipeline should only take the pre-trained model $\Theta_0$ and an unlabeled dataset $\mathbb{S}_{A}$ ($\{{x_{i}}\}^{N}_{i=1}$) accumulated from the specific local environment as input. We will then run the above graph-based meta-clustering module to obtain the estimated number of pseudo ``identity"(or class) labels as well as their belonging images. Let us denote $(x_i,y_i)$ as one training image embedding with its associated pseudo ID: $y_i$, one can then employ any of the recent loss functions such as ArcFace~\cite{ArcFace19}, AM-Softmax~\cite{Wang_2018_amsoftmax} and CircleLoss~\cite{CircleLoss} to train the adapted model $\Theta_{A}$ from $\Theta_0$.
\subsection{Preliminary of Face Recognition Training }

The early idea behind face recognition is to treat it as a classification problem and apply standard softmax loss to train the deep face representation: $\mathcal{L} = -\frac{1}{N} \sum_{i=1}^N \log \frac{e^{W^T_{{y_i}}f_i+b_{y_i}}}{\sum^{\mathcal{C}}_{j=1}e^{W^T_{j}f_i+b_i}}$
,where $f_i\in \mathbb{R}^d$ denotes the deep feature learned through the network for each image; $W_i \in \mathbb{R}^d$ represents the corresponding classifier vector from the last fully connected layer for label ${y_i}$, and $b_i$ is the bias term. By normalizing both classifier vector $\parallel W_j \parallel = 1$ and $\parallel f_i \parallel = 1$, while adding a scale factor $\gamma$, we get $W_j^Tf_i=\parallel w_j\parallel\parallel f_i\parallel\cos\theta_{j} = \cos\theta_j$, where $\theta_j$ is the angle between $w_j$ and $f_i$, for simplicity, we also fix $b_j=0$. To improve the generalization, recent study shows that adding different types of margins\cite{CircleLoss}\cite{Wang_2018_amsoftmax}\cite{ArcFace19} can bring significant gain in many places. For example, ArcFace\cite{ArcFace19} adds angular margin while AM-softmax \cite{Wang_2018_amsoftmax} adds cosine margin. Without loss of generality, we take AM-softmax to as the training objective: $ \mathcal{L}(\Theta) = - \frac{1}{N}\sum_{i=1}^N \log \frac{e^{\gamma(\cos({\theta_{y_i}) - m})}}{e^{\gamma(\cos({\theta_{y_i})- m})} + \sum^{\mathcal{C}}_{j \neq y_i} e^{\gamma\cos\theta_j}}$,
where $\mathcal{C}$ is the total number of classes, and $m$ is the margin that needs empirically determined.

\subsection{Regularized Center transfer (RCT)}
We assume pre-trained representation (from $\Theta_0$) already has strong discriminative power. After training, images belonging to the same identity tend to be clustered in a small region on the hyper-sphere, while still maintaining a cosine margin (defined as $m$) with images from different identities. If the pre-trained model was trained with a large number of different identities, the cluster area for each identity should have been squeezed on the hyper-sphere to reduce the intra-class variation. Let us denote $C^{\Theta_0}_{y_i} $ as the center of the face embedding for $y_i$ on the pre-trained model, then $ C^{\Theta_0}_{y_i} = \frac{1}{M_i}\sum^{M_i}_{k=1} \mathbbm{1}(y_k=y_i)f^{\Theta_0}_k$,
where $M_i$ is the total number of images belonging to class $y_i$. Inspired by center loss\cite{centerloss}, instead of learning $f_i$ and its corresponding $W_{y_i}$ from the new dataset from scratch, we propose to transfer this pre-trained class center as prior knowledge to the adapted model $\Theta_{A}$, so basically we want to have ${C_{y_i}^{\Theta_{A}}}$ to be as close  to  $C^{\Theta_0}_{y_i}$ as possible during the adaptation. In the context of AM-softmax\cite{Wang_2018_amsoftmax} or other latest training objectives such as CircleLoss \cite{CircleLoss}, we notice that classifier vector $W_{y_i}$ is also close to $C_{y_i}^{\Theta_0}$ even though they are not exactly the same. Therefore, in practice, to simplify the learning process, we directly use $C_{y_i}^{\Theta_0}$ to initialize each corresponding $W_{y_i}$, rather than learning it from scratch. To further reduce the overfitting risk when adapting to a small dataset, we add another model regularization term to let $\Theta_{A}$ not deviate too much from $\Theta_{0}$. Our final loss function is defined as follows:
\begin{align}
\label{eqn:RCT_Loss}
\begin{split}
        \mathcal{L}(\Theta_{A}) = {}&- \frac{1}{N}\sum_{i=1}^N \log \frac{e^{\gamma(\cos({\theta_{y_i}) -  m})}}{e^{\gamma(\cos({\theta_{y_i}) - m})} + \sum^{\mathcal{C}}_{j \neq y_i} e^{\gamma\cos\theta_j}} \\
      &+ \lambda \parallel \Theta_{A} - \Theta_{0} \parallel^2_2
\end{split}
\end{align}
subject to
\begin{align}
\begin{split}
     {}&W = {W^*}/{\parallel W^* \parallel}, \\
     &f = {f^*}/{\parallel f^* \parallel}, \\
     &\cos\theta_j= W^T_j f_i ,\\
     & W_{y_i} = C^{\Theta_0}_{y_i}
\end{split}
\end{align}
where $W$ is the normalized classifier matrix, $x$ is the normalized feature vector for each image,  $\lambda$ is a hyper-parameter to trade-off between the loss on the new dataset and the model regularization term. During the model adaptation training, we initialize each $W_{y_i}$ with $C^{\Theta_0}_{y_i}$ and keep it fixed and only fine-tune the feature representation. Compared with standard transfer learning without RCT, our regularized center transfer can better preserve the pre-trained representation, especially for small datasets, therefore reducing the risk of overfitting.

\subsection{Federated Aggregation}
As we argued in Sec.~\ref{sec.intro}, with the rising concern of data privacy, it is reasonable to assume that training data would be highly de-centralized across different clients in the future, where model adaptation is conducted first locally at each client and later on being aggregated via federated learning~\cite{FederatedAVg} in a secure way. We argue that our proposed unsupervised modal adaptation is designed to further facilitate such fully automated ``dual-loop" learning paradigm. To reduce the risk of losing privacy and adversarial attack, we remove the top classification layer and only transfer the backbone model parameters between each client and server and conduct simple model averaging, we denote such partial averaging setup as ``FedPav" in our experiments \cite{Zhuang_2020}.

\section{Experimental Settings}
\subsection{Datasets and Protocols}
In each Local-Adaptive face recognition (LaFR) protocol, the pre-trained face recognition model is trained with base dataset $\mathbb{S}_{B} = \{{x^{B}_{i}},{y^{B}_{i}}\}$, and the meta-cluster GCN model is trained with multiple non-overlapped labeled datasets $\mathbb{S}_{C} = \{{x^{{C}_{1}}_{i}},{y^{{C}_{1}}_{i}}\} \cup \{{x^{{C}_{2}}_{i}},{y^{{C}_{2}}_{i}}\} \cup ... \{{x^{{C}_{k}}_{i}},{y^{{C}_{k}}_{i}}\}$. While deploying the model in the specific scenario, the model is adapted with unlabeled dataset $\mathbb{S}_{A} = \{{x^{A}_{i}}\}$ from the environment. The final face recognition performance is evaluated on the testing dataset $\mathbb{S}_{T} = \{{x^{T}_{i}},{y^{T}_{i}}\}$. Here, $x_i$ and $y_i$ represent the i-th face image and the corresponding identity label, respectively. Note that the base dataset $\mathbb{S}_{B}$ is not available during clustering and adaptation. 
\vspace{-3mm}

\paragraph{Adapt to Different Local Races} The existence of face image distribution shift between different races has been proved in related works \cite{Wang_2019_ICCV}\cite{wang2020mitigating}, so we re-organized the racial datasets collected by \cite{Wang_2019_ICCV} to build the LaFR protocols. We leverage the Caucasian dataset from BUPT-Transferface~\cite{Wang_2019_ICCV} as $\mathbb{S}_{B}$ for base face recognition model training, and racial faces in-the-wild (RFW) dataset~\cite{Wang_2019_ICCV} is used as $\mathbb{S}_{T}$ for testing after adaptation. BUPT-Balancedface~\cite{Wang_2019_ICCV}, which has non-overlapped 7k subjects for each race, is leveraged for meta-cluster model training and further adaptation. The dataset details are summarized in Table~\ref{tab:dataset1}. We build three leave-one-out protocols, which one of the races from \{African, Asian, Indian\} is selected as the local target, and dataset from other races can be used for meta-cluster GCN model training.
\vspace{-3mm}

\begin{figure}[t]
    \centering
    \includegraphics[width=0.8\linewidth]{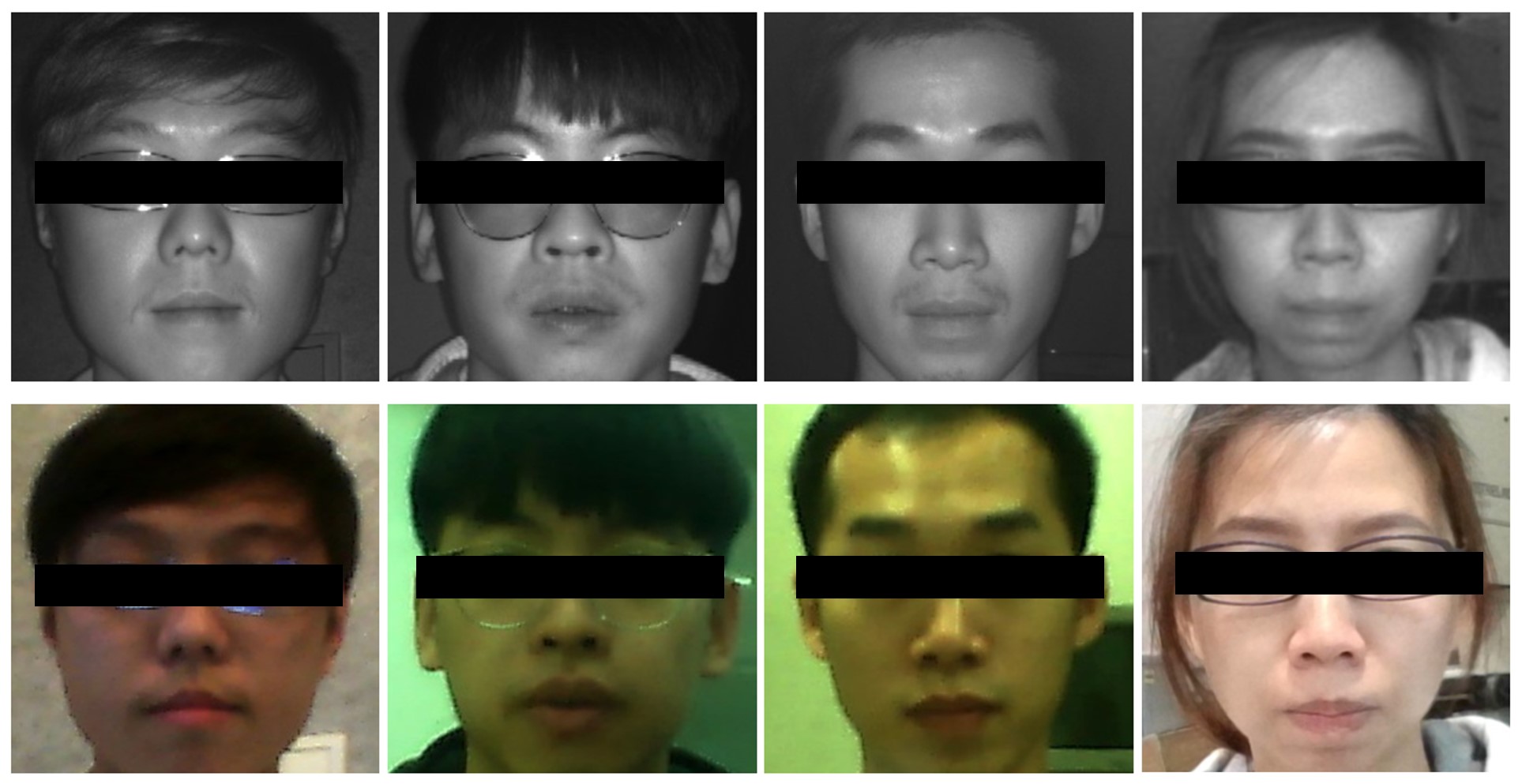}
    \caption{\textbf{Top:} Infrared Images captured by 4 different infrared cameras (the first two capture 940nm wavelength, the other two 850nm); \textbf{Bottom:} Their corresponding RGB images captured by another 4 RGB cameras.}
    \label{fig:sensors}
    \vspace{-4mm}
\end{figure}

\paragraph{Adapt to Different Local Infrared Sensors}
Different wavelengths have different penetration rates into human skins, leading to different contrasts on the IR face images. Most cameras are sensitive to 850nm infrared light and generally considered to have higher contrast than 940nm infrared light. Meanwhile, different cameras also adapt different Image Signal Processors (ISP), so the way to handle tone-mapping, noise reduction, and blurriness are different too, which causes another type of significant appearance distribution bias across different sensors. In this paper, we are interested in studying how to adapt a pre-trained RGB face recognition model to different local environments, each of which mimics a specific infrared camera enabled face recognition scenario. As there is no public infrared camera dataset with different ISP or wavelength, we collected an internal dataset with four different infrared camera sensors that capture infrared wavelength ranging from 850 nm to 940 nm. Figure \ref{fig:sensors} shows a few typical examples sampled from our four datasets, whose details are summarized in Table~\ref{tab:dataset2}. We partition the collected dataset by identity to form $\mathbb{S}_{C}$, $\mathbb{S}_{A}$, and $\mathbb{S}_{T}$, and we leverage commonly used RGB dataset MS-1M~\cite{guo2016msceleb1m} as $\mathbb{S}_{B}$. Similarly, we build four leave-one-out protocols, where we use datasets from 3 sensors to train our proposed meta-clustering while the remaining one for both adaptation and final testing. 

\begin{table}[t]
    \centering
    \resizebox{0.7\columnwidth}{!}{\begin{tabular}{c|ccc}
        \hline
        Dataset & Race & \# Subjects  & \# Images \\
        \hline
        \hline
        $\mathbb{S}_{B}$ & Caucasian & 10000 & 468139 \\
        \hline
        \multirow{3}{*}{$\mathbb{S}_{C}, \mathbb{S}_{A}$} & Indian & 7000 & 275095 \\
        & Asian & 7000 & 325475 \\
        & African & 7000 & 324376 \\
        \hline
        \multirow{3}{*}{$\mathbb{S}_{T}$} & Indian & 2984 & 10308 \\
        & Asian & 2492 & 9688 \\
        & African & 2995 & 10415 \\
        \hline
    \end{tabular}}
    \caption{Statistics of the datasets for races.}
    \label{tab:dataset1}
    \vspace{-3.5mm}
\end{table}

\begin{table}[t]
    \centering
    \resizebox{0.7\columnwidth}{!}{\begin{tabular}{c|cccc}
        \hline
        Dataset & Sensor & \# Subjects  & \# Images \\
        \hline
        \hline
        $\mathbb{S}_{B}$ & RGB & 94,430 & 5,179,510 \\
        \hline
        \multirow{4}{*}{$\mathbb{S}_{C}, \mathbb{S}_{A}$} 
        & IR-A & 430 & 57,818 \\
        & IR-B & 250 & 34,592 \\
        & IR-C & 220 & 30,689 \\
        & IR-D & 400 & 55,312 \\
        \hline
        \multirow{4}{*}{$\mathbb{S}_{T}$} 
        & IR-A & 210 & 28,916 \\
        & IR-B & 3,372 & 27,155 \\
        & IR-C & 200 & 14,035 \\
        & IR-D & 235 & 32,388 \\
        \hline
    \end{tabular}}
    \caption{Statistics of the datasets for sensors.}
    \label{tab:dataset2}
    \vspace{-3.5mm}
\end{table}

\subsection{Evaluation Metrics} \label{metrics}
\paragraph{Face Embedding Clustering} Given the ground truth label from adaptation datasets $\mathbb{S}_{A}$, the intermediate face clustering result from the GCN model can be evaluated to indicate the accuracy of identity label assignment during adaptation. Two common clustering metrics~\cite{yang2019learning, yang2020learning}: \textit{Pairwise F-score} ($F_P$) and \textit{B-Cubed F-score} ($F_B$) are used, which calculate the harmonic mean of precision and recall. The metric $F_p$ puts relatively more emphasis on large face clusters, while $F_B$ weights clusters linearly based on their size.

\paragraph{Local Adaptive Face Recognition} The face recognition performance in the adaptation target is evaluated with standard face recognition metrics. In race adaptation protocols, we calculate the verification accuracy on 6000 difficult pairs selected by~\cite{Wang_2019_ICCV}. In sensor adaptation protocols, we report the False Non-Match Rates where the False Match Rate is 1e-6 (FNMR@FMR=1e-6).

\subsection{Implementation Details}
\paragraph{Meta-Clustering with GCN Training} For GCN model training, we set the meta learning rate $\alpha$ to 0.1, the outer loop learning rate $\beta$ to 0.1, and the meta loss weight $\xi$ to 1.0 to conduct the meta optimization. In all experiments, we set the momentum to 0.9 and train 30000 iterations with the SGD optimizer. In each iteration, we randomly select one dataset from $\mathbb{S}_{C}$ as the meta-test set and the other datasets as the meta-train set. It takes around 7 hours to train on one TITAN X GPU.
\vspace{-3mm}

\paragraph{Model Adaptation} After clustering, we follow ~\cite{yang2020learning} to set a threshold $\tau$ to 0.8 to cut off the edges with small similarities, and obtain the pseudo identity labels by simply connecting each vertex to their nearest neighbors. During model adaptation with Regularized Center Transfer (RCT), we follow Equ. \ref{eqn:RCT_Loss} to calculate the face embeddings center of each identity according to the pseudo labels. We set the model regularization $\lambda$ to 0.1 and employ CircleLoss~\cite{CircleLoss} as the classification loss. In all adaptation experiments, we set the learning rate to 0.001 and train for 50 epochs. It takes around 5 hours to train on one TITAN X GPU.

\section{Experimental Results}
\subsection{Adapt to Different Races}
\paragraph{Face Embedding Clustering}
With the ground truth identity label (not available during clustering and adaptation) from the dataset $\mathbb{S}_{A}$, we can evaluate the intermediate face clustering results from different methods using the F-score metrics described in Sec. \ref{metrics}. We compare the clustering performance with the original GCN~\cite{yang2020learning} method, which uses all the datasets in $\mathbb{S}_{C}$ to build a large graph and learn the embedding confidence prediction through the supervised training scheme described in \cite{yang2020learning}. Besides, we also compare with another distance-based clustering method~\cite{Wang_2019_ICCV}, which connects all the embedding pairs whose cosine distance between them is less than a fixed threshold. We report the result with the threshold 0.3, which has the best performance across all sampled thresholds. The clustering performance of racial adaptation protocols is shown in Table \ref{tab:cluster1}, and we denote our proposed meta-clustering method as ``meta-GCN". From the results, we can observe that the simple distance-based clustering method~\cite{Wang_2019_ICCV} has a very low Pairwise F-score ($F_P$) but high Bcubed F-score ($F_B$) in some adaptation protocols, which means that it cannot handle large embedding clusters. Our proposed ``meta-GCN" outperforms the original GCN~\cite{yang2020learning} in most of the benchmarks. It indicates that the proposed meta-learning scheme can learn more generalized GCN parameters and cluster the face embeddings better in unseen local environments. The clustering output with higher F-scores produces cleaner pseudo identity labels, which benefit more to the model adaptation process.
\vspace{-3mm}

\begin{table}[t]
    \centering
    \resizebox{0.95\columnwidth}{!}{\begin{tabular}{c|ccc}
        \hline
        Methods & African & Asian & Indian \\ \hline \hline
        Distance-based~\cite{Wang_2019_ICCV} & 0.0086 / 0.7282 & 0.0035 / 0.6267 & 0.6891 / 0.7481 \\
        GCN~\cite{yang2020learning} & 0.6115 / 0.8132 & 0.3492 / 0.6432 & 0.8559 / \textbf{0.8725}\\
        Meta-GCN & \textbf{0.8129} / \textbf{0.8535} & \textbf{0.3768} / \textbf{0.6876} & \textbf{0.8849} / 0.8551 \\
        \hline
    \end{tabular}}
    \caption{Comparison of face embedding clustering performance on three racial adaptation protocols. Two common clustering metrics: Pairwise / Bcubed F-score ($F_P / F_B$) pairs are reported.}
    \label{tab:cluster1}
    \vspace{-3.5mm}
\end{table}

\begin{figure}[t!]
    \centering
    \includegraphics[width=.7\linewidth]{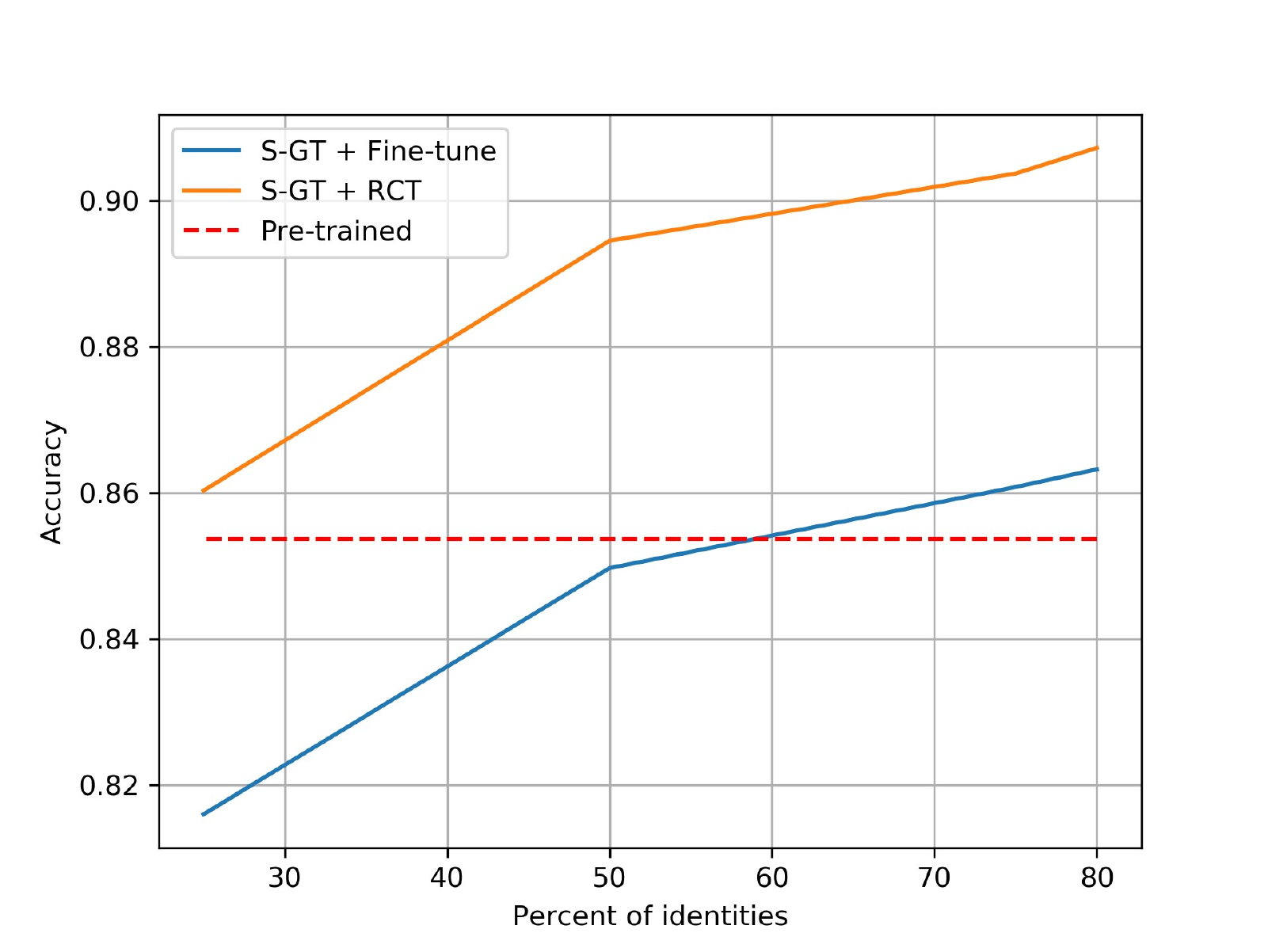}
    \caption{Performance comparison between \textbf{$\mathbb{S}_{A}$-GT + Fine-tune} and \textbf{$\mathbb{S}_{A}$-GT + RCT} under different percent of identities on the ``African" adaptation protocol.}
    \label{fig:ACC_DATASET_SIZE}
    \vspace{-3.5mm}
\end{figure}

\paragraph{RCT Technique for Model Adaptation}
To evaluate the effectiveness of our proposed Regularized Center Transfer (RCT) technique, we leverage the ground truth (GT) identity labels from the adaptation dataset $\mathbb{S}_{A}$ to perform model adaptation. We compare the result with standard transfer learning, which is adapting the face recognition model without regularizing the classifier during the adaptation process. From Table~\ref{tab:GCNV_ADAPT}, we have the following observations. First, the pre-trained model cannot perform well on three local targets without adaptation due to the race gap. Second, standard fine-tuning could possibly harm the face recognition model (ex. Indian protocol) due to the risk of overfitting on small dataset, which is illustrated in Figure \ref{fig:ACC_DATASET_SIZE}) where it shows RCT is superior to standard fine-tuning across different size of identities. Third, our regularization techniques can effectively prevent the local model from overfitting and optimize the representation better than fine-tuning by a large margin in all protocols. Furthermore, we conduct ablation experiments to study the effect of different loss functions and the regularization term used in RCT. Results from Table~\ref{tab:ACC1} demonstrate that: 1) Our RCT can work with other margin-based classification loss like AM-softmax~\cite{Wang_2018_amsoftmax}, but perform better with the SOTA CircleLoss~\cite{CircleLoss}. 2) The weight regularization term (controlled by $\lambda$) does contribute a lot to the final adaptation performance.   
\vspace{-3mm}

\begin{table}[t]
    \centering
    \resizebox{0.85\columnwidth}{!}{\begin{tabular}{c|cccc}
        \hline
        Methods & Caucasian & African & Asian & Indian \\ 
        \hline 
        \hline
        Pre-trained & 0.9512 & 0.8537 & 0.8633 & 0.9047 \\
        $\mathbb{S}_{A}$-GT + Fine-tune & - & 0.8758 & 0.8705 & 0.8528 \\ 
        $\mathbb{S}_{A}$-GT + RCT & - & \textbf{0.9227} & \textbf{0.9212} & \textbf{0.9323} \\
        \hline
    \end{tabular}}
    \caption{Deployment model performance comparison while using the ground-truth (GT) labels from adaptation dataset ($\mathbb{S}_{A}$) on three racial protocols. Verification accuracy on 6000 pairs of the testing dataset ($\mathbb{S}_{T}$) is reported.}
    \label{tab:GCNV_ADAPT}
    \vspace{-3.5mm}
\end{table}
\begin{table}[t]
    \centering
    \resizebox{0.85\columnwidth}{!}{\begin{tabular}{c|cccc}
        \hline
        Methods & African & Asian & Indian \\ \hline \hline
        Pre-trained & 0.8537 & 0.8633 & 0.9047 \\
        \hline
        RCT (AM-Softmax, $\lambda$=0.0) & 0.9125 & 0.9050 & 0.9048 \\
        RCT (CircleLoss, $\lambda$=0.0) & 0.9145 & 0.9033 & 0.9120 \\
        RCT (AM-Softmax, $\lambda$=0.1) & 0.9158 & 0.9153 & 0.9278 \\
        RCT (CircleLoss, $\lambda$=0.1) & \textbf{0.9227} & \textbf{0.9212} & \textbf{0.9323} \\
        \hline
    \end{tabular}}
    \caption{The influence of loss function choices and weight regularization $\lambda$ on racial adaptation protocols. All experiments are conducted with ``$\mathbb{S}_{A}$-GT + RCT" setting.}
    \label{tab:ACC1}
    \vspace{-3.5mm}
\end{table}

\paragraph{End-to-end Unsupervised LaFR}
We further conduct end-to-end Local-Adaptive face recognition (LaFR) experiments, which combine face embedding clustering methods (from Table~\ref{tab:cluster1}) with the proposed RCT adaptation technique. The end-to-end LaFR performance on three race protocols are shown in Table~\ref{tab:end-to-end}. The result in the last row which adopts ground truth (GT) labels from $\mathbb{S}_{A}$ serves as the upper bound of end-to-end unsupervised LaFR methods. We also re-implement and compare with the recent generalized face recognition method MFR \cite{guo2020learning} in our protocol. The dataset in $\mathbb{S}_{C}$ are leveraged into the meta-optimization process to obtain a better universal model. Although MFR works better than standard pretraining,  we still observe the performance gap between MFR and our adaptation results, indicating the value of local adaptation for optimized performance. Meanwhile, Table~\ref{tab:end-to-end} also shows that our proposed ``meta-GCN" outperforms both distance-based\cite{Wang_2019_ICCV} and regular GCN\cite{yang2020learning} based clustering methods in our protocols, which proves the effectiveness of the graph representation as well as the meta learning on top of the GCN. 

\begin{table}[t]
    \centering
    \resizebox{0.95\columnwidth}{!}{\begin{tabular}{c|cccc}
        \hline
        Methods & African & Asian & Indian \\ 
        \hline 
        \hline
        Pre-trained & 0.8537 & 0.8633 & 0.9047 \\
        \hline
        MFR~\cite{guo2020learning} & 0.8882 & 0.8768 & 0.9118 \\
        \hline
        Distance-based~\cite{Wang_2019_ICCV} + RCT & 0.7613 & 0.7488 & 0.8862 \\
        GCN~\cite{yang2020learning} + RCT & 0.8692 & 0.8717 & 0.8940 \\
        Meta-GCN + RCT & \textbf{0.8912} & \textbf{0.8773} & \textbf{0.9258} \\
        \hline
        $\mathbb{S}_{A}$-GT + RCT & 0.9227 & 0.9212 & 0.9323 \\
        \hline
    \end{tabular}}
    \caption{Deployment model performance comparison while using pseudo identity labels from \textbf{different clustering methods} combined with RCT on three racial protocols. Verification accuracy on 6000 pairs of the testing dataset ($\mathbb{S}_{T}$) is reported.}
    \label{tab:end-to-end}
    \vspace{-3.5mm}
\end{table}

\begin{figure}[t!]
    \centering
    \includegraphics[width=.9\linewidth]{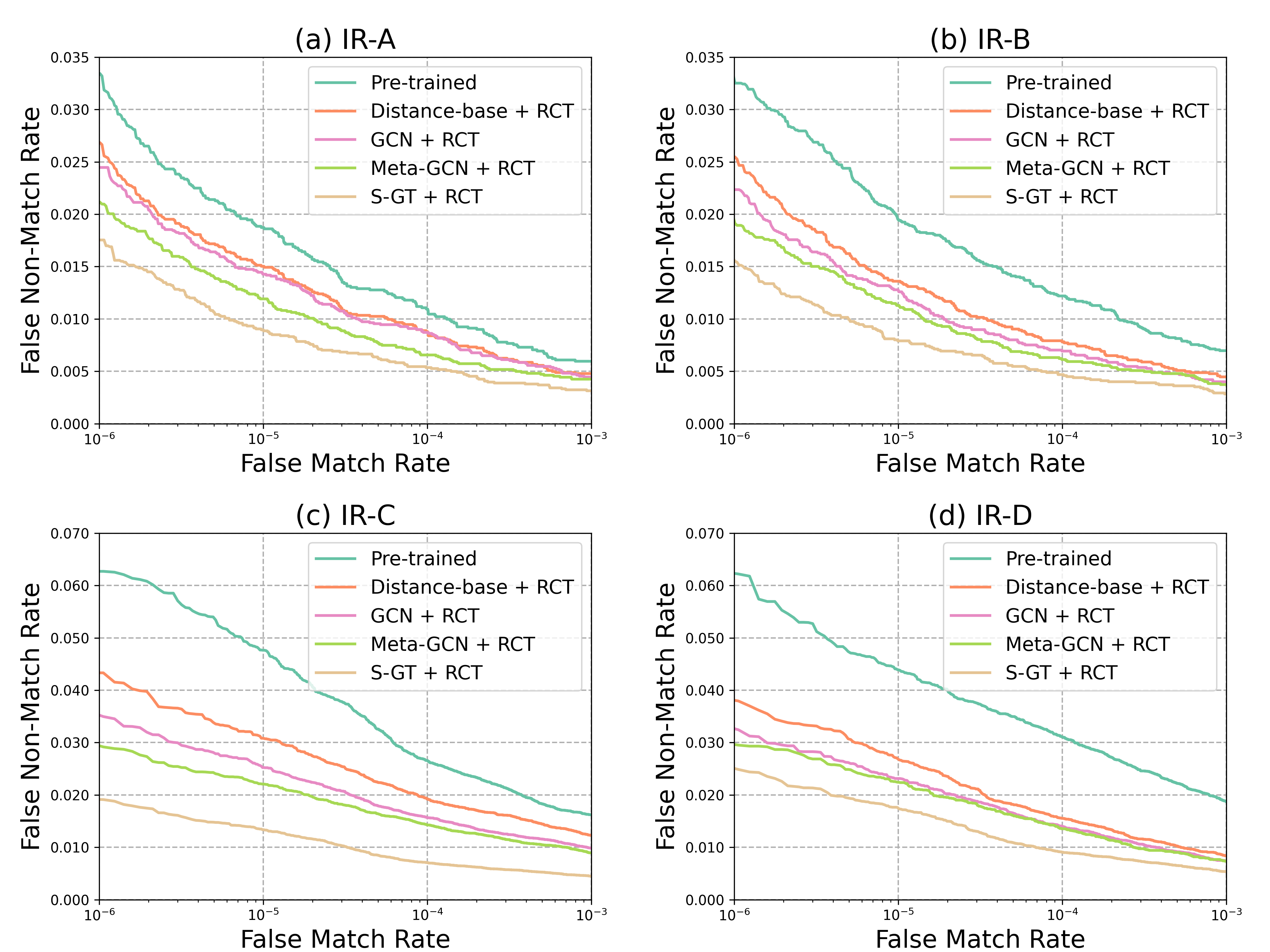}
    \caption{The ROC curves of deployed models evaluated on all pairs of (a) IR-A, (b) IR-B, (c) IR-C, (d) IR-D protocols.}
    \label{fig:roc2}
\end{figure}

\begin{table}[t]
    \centering
    \bgroup
    \def\arraystretch{1.0}
    \setlength\tabcolsep{3.0pt}
    \resizebox{0.95\columnwidth}{!}{\begin{tabular}{ccc|cccc}
        \hline
        Label & Loss & FedPav & IR-A & IR-B & IR-C & IR-D \\ 
        \hline 
        \hline
        $\mathbb{S}_{B}$-GT & (Pre-trained) &  & 0.0344 & 0.0330 & 0.0633 & 0.0626 \\
        \hline
        SpC~\cite{ge2020selfpaced} & UCL~\cite{ge2020selfpaced} &  & 0.0332 & 0.0298 & 0.0579 & 0.0623 \\ 
        SpC~\cite{ge2020selfpaced} & RCT~ & & 0.0256 & 0.0256 & 0.0443 & 0.0425  \\ 
        D-based~\cite{Wang_2019_ICCV} & RCT & & 0.0267 & 0.0254 & 0.0433 & 0.0384 \\
        GCN~\cite{yang2020learning} & RCT & & 0.0245 & 0.0223 & 0.0358 & 0.0331 \\
        Meta-GCN & RCT & & 0.0212 & 0.0198 & 0.0298 & 0.0299  \\
        Meta-GCN & RCT & \checkmark & \textbf{0.0207} & \textbf{0.0188} & \textbf{0.0274} & \textbf{0.0267}  \\
        \hline
        $\mathbb{S}_{A}$-GT & RCT & & \underline{0.0175} & \underline{0.0158} & \underline{0.0192} & \underline{0.0251} \\
        $\mathbb{S}_{A}$-GT & RCT & \checkmark & \underline{0.0175} & \underline{0.0164} & \underline{0.0187} & \underline{0.0228}  \\
        \hline
    \end{tabular}}
    \egroup
    \caption{Comparison between different LaFR methods on sensor adaptation protocols. The performance is measured by FNMR@FMR=1e-6, the lower the better.}
    \label{tab:GCNV_ADAPT_IR}
\end{table}

\subsection{Adapt to Different Sensors}
We conduct similar end-to-end LaFR experiments on four infrared sensor adaptation protocols, note that those sensors are from real industry face recognition systems. From Table~\ref{tab:GCNV_ADAPT_IR} and Figure~\ref{fig:roc2}, we have the following observations. First, as the upper bound for tested end-to-end LaFR methods, ``$\mathbb{S}_{A}$-GT + RCT" can reduce the adaptation non-match error up to 70\% compared with the strong RGB pre-trained model, which demonstrate the effectiveness of our adaptation regularization techniques for face recognition. Second, with limited number of identities collected in the environment (ex. 220 in IR-C), standard fine-tuning (``$\mathbb{S}_{A}$-GT + fine-tune") is prone to overfitting and cannot achieve better performance while sensor adaptation compared with the pre-trained model. Third, our proposed ``Meta-GCN + RCT" achieves the best performance in all protocols among different clustering methods, which shows the generalization ability and robustness of our meta-clustering GCN model in different scenarios. 
\paragraph{Compared with SOTA Domain-adaptive ReID} To show the necessity of our proposed solution, we further compare with the SOTA object Re-ID method \cite{ge2020selfpaced}. We applied both their Self-paced Clustering(SpC) and the Unified Contrastive Loss(UCL) in our system. Even though SpC performs impressively on person reID task, their techniques are still inferior to our proposed Meta-GCN with RCT on the task of sensor adaptation for face recognition, which is likely caused by the unique challenging nature of face recognition task.
\vspace{-3mm}
\paragraph{Experiments on Federated Learning} We conduct experiments to verify the effectiveness of our LaFR in federated learning setting. Assuming there are $K$ clients ($i=1,2,...,K$), each associates with unlabeled private dataset $D_{i}$, and the face recognition model $\Theta_{i}^{t}$ at the optimization dual-loop step $t$. The continual model adaptation process within federated learning setup is shown in Algorithm~\ref{alg:fl}. The training procedure contains 20 rounds of dual-loop, and each dual-loop consists of a local adaptation and a partial federated aggregation (FedPav) which averages the backbone parameters from the local clients. The results shown in Tab.~\ref{tab:GCNV_ADAPT_IR} verify the effectiveness of the federated learning pipeline to iteratively improve the local face recognition models without breaching user's privacy on each client.
\vspace{-1.5mm}
\begin{algorithm}[!tb]
\caption{Model Adaptation in Federated Learning}
\label{alg:fl}
\begin{algorithmic}[1]
\REQUIRE
 Pretrained model on server $\Theta_{s}^{0}$;
 Client models $\Theta_{c_{i}}^{0}$;
 Client datasets $D_{i}$ (i=1,2,...,K); 
\ENSURE Client models $\Theta_{c_{i}}^{T}$;\\
\FOR{each step t = 0 to T-1} 
\FOR{each model i = 1 to K} 
\STATE  $\Theta_{c_{i}}^{t+1}$ $\leftarrow$ RCT($\Theta_{s}^{t}$, $\Theta_{c_{i}}^{t}$, $D_{i}$)
\ENDFOR
\STATE $\Theta_{s}^{t+1}$ $\leftarrow$ $\frac{1}{K}\sum_{i=1}^{K}$ $\Theta_{c_{i}}^{t}$
\ENDFOR
\end{algorithmic}
\end{algorithm}

\section{Conclusions}
We introduce a new problem setup called ``Local-Adaptive Face Recognition (LaFR)", which aims to produce specialized face recognition models tailored for each local environment. Our proposed graph-based meta-clustering model can better cluster the face embeddings for unseen environments, which provides cleaner identity labels in an unsupervised manner during adaptation. Combined with the proposed RCT module, our framework can robustly produce LaFR models adapted from imperfect pre-trained face recognition models. The effectiveness of our framework is proven on various protocols, including racial and sensor adaptation, with or without federated aggregation. We hope these efforts can open up future directions towards specialized face recognition models at scale.

{\small
\bibliographystyle{ieee_fullname}
\bibliography{egbib}
}

\end{document}